\documentclass[sigplan,screen,nonacm]{acmart}
\AtBeginDocument{%
  \providecommand\BibTeX{{%
    \normalfont B\kern-0.5em{\scshape i\kern-0.25em b}\kern-0.8em\TeX}}}

\begin{document}

\title{Multimodality of AI for Education: Towards Artificial General Intelligence}

\author{Gyeong-Geon Lee}
\affiliation{%
  \institution{AI4STEM Education Center, University of Georgia}
  \city{Athens}
  \state{GA}
  \country{USA}
}
% \email{gyeong-geon.lee@uga.edu} % Assuming email

\author{Lehong Shi}
\affiliation{%
  \institution{Department of Workforce Education and Instructional Technology, University of Georgia}
  \city{Athens}
  \state{GA}
  \country{USA}
}
% \email{lehong.shi@uga.edu} % Assuming email

\author{Ehsan Latif}
\affiliation{%
  \institution{AI4STEM Education Center, University of Georgia}
  \city{Athens}
  \state{GA}
  \country{USA}
}
% \email{ehsan.latif@uga.edu} % Assuming email

\author{Yizhu Gao}
\affiliation{%
  \institution{Department of Mathematics, Science, and Social Studies Education, University of Georgia}
  \city{Athens}
  \state{GA}
  \country{USA}
}
% \email{yizhu.gao@uga.edu} % Assuming email

\author{Arne Bewersdorff}
\affiliation{%
  \institution{Department of Educational Sciences, School of Social Sciences and Technology, Technical University of Munich}
  \city{Munich}
  \state{BY}
  \country{Germany}
}
% \email{arne.bewersdorff@tum.de} % Assuming email

\author{Matthew Nyaaba}
\affiliation{%
  \institution{AI4STEM Education Center \& Department of Educational Theory and Practice, University of Georgia}
  \city{Athens}
  \state{GA}
  \country{USA}
}
% \email{matthew.nyaaba@uga.edu} % Assuming email

\author{Shuchen Guo}
\affiliation{%
  \institution{College of Teacher Education, Nanjing Normal University}
  \city{Nanjing}
  \state{Jiangsu}
  \country{China}
}
% \email{gsc44@nnu.edu.cn} % Assuming email

\author{Zihao Wu}
\affiliation{%
  \institution{School of Computing,\\ University of Georgia}
  \city{Athens}
  \state{GA}
  \country{USA}
}
% \email{zihao.wu1@uga.edu} % Assuming email

\author{Zhengliang Liu}
\affiliation{%
  \institution{School of Computing,\\ University of Georgia}
  \city{Athens}
  \state{GA}
  \country{USA}
}
% \email{zl18864@uga.edu}

\author{Hui Wang}
\affiliation{%
  \institution{Second Language Acquisition and Teaching, University of Arizona}
  \city{Tucson}
  \state{AZ}
  \country{USA}
}
% \email{hwang0524@arizona.edu} % Assuming email

\author{Gengchen Mai}
\affiliation{%
  \institution{Department of Geography \& School of Computing, University of Georgia}
  \city{Athens}
  \state{GA}
  \country{USA}
}
% \email{gengchen.mai@uga.edu} % Assuming email

\author{Tiaming Liu}
\affiliation{%
  \institution{AI4STEM Education Center \& School of Computing, University of Georgia}
  \city{Athens}
  \state{GA}
  \country{USA}
}
% \email{tiaming.liu@uga.edu} % Assuming email

\author{Xiaoming Zhai}
\authornote{Corresponding author. Email: xiaoming.zhai@uga.edu}
\affiliation{%
  \institution{AI4STEM Education Center \& Department of Mathematics, Science, and Social Studies Education, University of Georgia}
  \city{Athens}
  \state{GA}
  \country{USA}
}
% \email{}

\renewcommand{\shortauthors}{Lee and Shi, et al.}

\begin{abstract}
    This paper presents a comprehensive examination of how multimodal artificial intelligence (AI) approaches are paving the way towards the realization of Artificial General Intelligence (AGI) in educational contexts. It scrutinizes the evolution and integration of AI in educational systems, emphasizing the crucial role of multimodality, which encompasses auditory, visual, kinesthetic, and linguistic modes of learning. This research delves deeply into the key facets of AGI, including cognitive frameworks, advanced knowledge representation, adaptive learning mechanisms, strategic planning, sophisticated language processing, and the integration of diverse multimodal data sources. It critically assesses AGI's transformative potential in reshaping educational paradigms, focusing on enhancing teaching and learning effectiveness, filling gaps in existing methodologies, and addressing ethical considerations and responsible usage of AGI in educational settings. The paper also discusses the implications of multimodal AI's role in education, offering insights into future directions and challenges in AGI development. This exploration aims to provide a nuanced understanding of the intersection between AI, multimodality, and education, setting a foundation for future research and development in AGI.
\end{abstract}






\maketitle

\begin{figure}[htp]
\begin{center}
\includegraphics[width=1.0\linewidth]{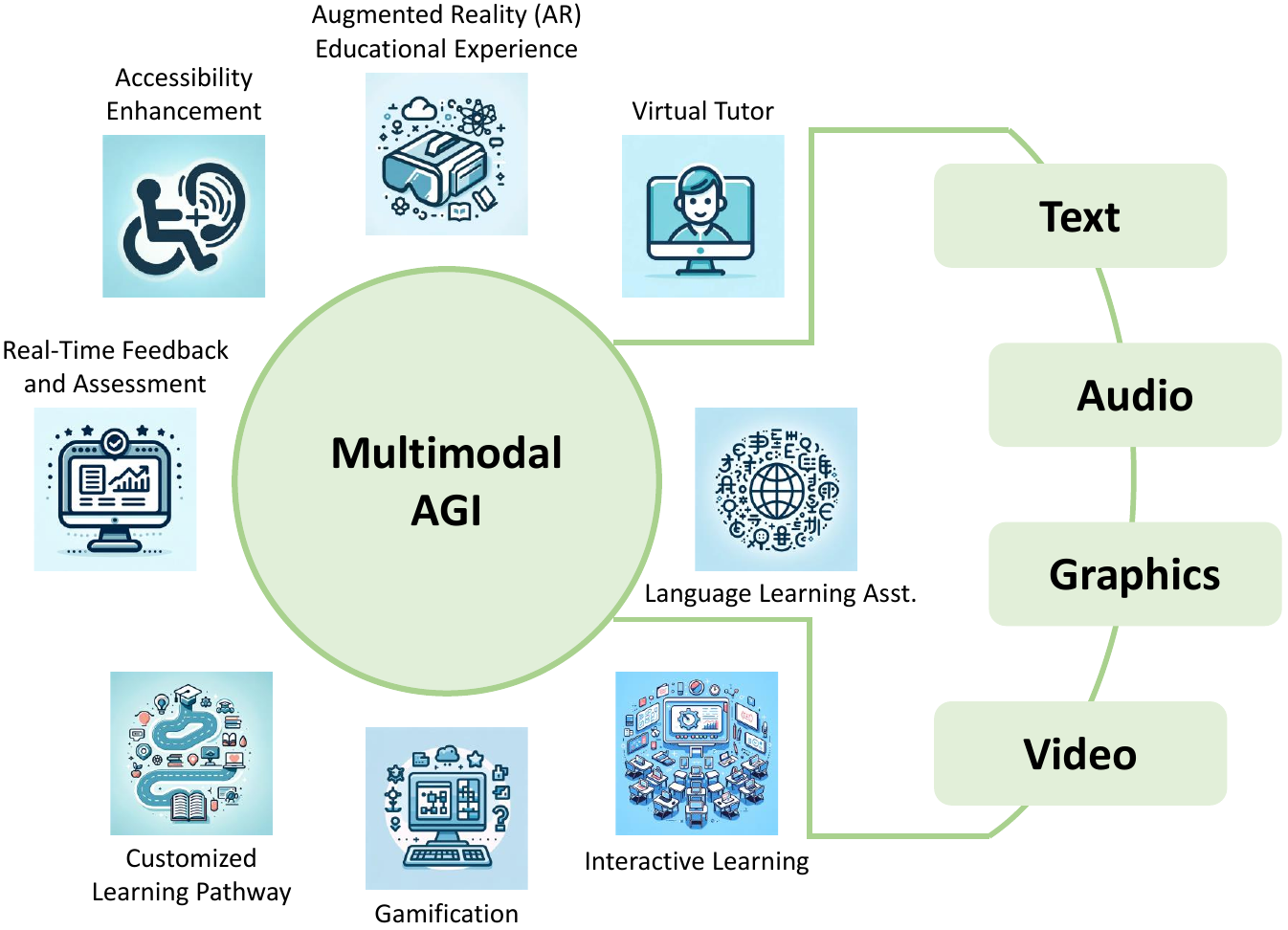}
\end{center}
\caption{Overview of Multimodal AGI for Education.} 
\label{fig:overview}
\end{figure}

\section{Introduction}
Integrating Artificial Intelligence (AI) in education has become a transformative force as the educational landscape continually evolves. Inspired by the recent development of GPT-4V \citep{openai2023gpt4} and Gemini\citep{gemini2023}, the idea of multimodality in AI for education heralds a synergistic confluence of various sensory channels and cognitive strategies to enhance teaching-learning and assessment. This exploration commences with an exposition of the multimodal nature of learning — how learners engage with information via auditory, visual, and kinesthetic means \citep{mayer1994picture} — and the consequent imperative for educational technologies to cater to this diversity. With its capacity to process and synthesize vast streams of multimodal information, AI is a potent ally in this regard \citep{sharma2019building}. Particularly, it is evidenced that the multifaceted dimensions of multimodal AI applications cast light on the potential to pave the way toward Artificial General Intelligence (AGI) within the educational domain.  

AGI is a hypothetical machine able to understand, learn, and apply knowledge across a range of cognitive tasks at a human-like level \cite{mclean2023risks,torres2019possibility}. Recent progress in AI technologies showcases a wide spectrum of intelligence in text, graphic, video, and audio analysis and generation. Particularly, the progress in generative AI has set forth the mission for AGI as a significant stride for society. For example, Elon Musk is on a mission to create the world’s first AGI \citep{Das2023@Musk}. OpenAI claimed pioneering research on the path to AGI -- "We believe our research will eventually lead to artificial general intelligence [AGI], a system that can solve human-level problems. Building safe and beneficial AGI is our mission \citep{OpenAI@AGI}." Google also deployed Gemini, which reportedly outperforms human experts since it "is built from the ground up for multimodality - reasoning seamlessly across text, images, video, audio, and code" \citep{deepmind_gemini}. While AGI remains a nascent aspiration, the strides in multimodal AI in education provide a microcosm of the broader potentials and challenges inherent in pursuing such a comprehensive form of intelligence. Conceptualizing the integration of AGI into education paves the way for incoming groundbreaking innovations that are redefining traditional teaching and learning paradigms.

Through this paper, we aim to provide a conceptual framework for understanding the role of multimodal AI in educational settings and to speculate on how these advancements might inform and inspire the trajectory of AGI for education (see Fig.~\ref{fig:overview} for an overview if multimode AGI for education). By weaving together the threads of current AI capabilities with the tapestry of educational needs and outcomes, we propose a vision where AI supports, transforms, and redefines the pedagogical processes, embodying collaborative intelligence alongside educators and learners. In pursuit of this vision, we examine the current state of AI applications in education, identify the gaps and opportunities for more sophisticated multimodal interactions, and project the implications of these technologies for the development of AGI for education. Realizing that the journey toward AGI is fraught with technical challenges and ethical considerations, we discussed the issues and concerns, providing a fertile ground for addressing these issues in a controlled, impactful, and benevolent manner.

\section {Theoretical Foundations of Multimodality in Learning}\label{Gyeong-Geon-Yizhu}

\subsection{Concept of Multimodality } \label{Gyeong-Geon}

The learning and information processing channels in humans' minds are notably impactful across diverse educational activities. One of the most noteworthy manifestations is the fusion and synergy of multimodality in education \citep{kress2001multimodal}. Human cognitive processing of a construct is made through various information channels that process information in different forms, including text, image, sound, gaze, etc. These different types of information media are referred to as “modalities,” with each of them partially contributing to sense-making and playing a discrete role in the whole process of learning \citep{kress2001multimodal, fleming1992not}.

Researchers have realized the multi-modality feature of human learning for centuries. The early history could be traced back to John Amos Comenius's milestone work \textit{Orbis Pictus} (1st ed., 1658) \citep{comenius1887orbis, iverson1953audiovisual}. \textit{Orbis Pictus} is the first picture textbook in the history of education covering a wide range of subjects, such as inanimate nature, zoology, and Botanics. As a textbook read by European children for more than 200 years \citep{boyd1921history}, the content in most editions was presented in children's mother tongue, Latin language, and images, and the objects being explained in the two languages simultaneously. Orbis Pictus had a lasting impact on children's education, serving as an early model for both the visual and lexical approaches in early education.

In the 19th century, magic lantern slides emerged as learning media to convey information for learning, which served as the early version of the projector. This invention can visualize the structure, stories, etc., and concretes learning content and has been deemed a milestone for visual aids in education. In the early 20th century (1900-1950), motion pictures in the forms of film, television, or simulations could vividly present stories, which significantly increased the amount of information conveyed in the learning media and children's learning interest \citep{iverson1953audiovisual}. As audio-visual media was increasingly being adopted in the public domains, such as schools, museums, and libraries, particularly after World War II, there was a practical need to uncover how multi-media plays a role in the learning processes  \citep{freeman1990visualmedia}.

\subsection{Dual Coding Theory and Multimedia Theory} \label{Gyeong-Geon}
\begin{figure}[htp]
    \centering
    \includegraphics[width=\linewidth]{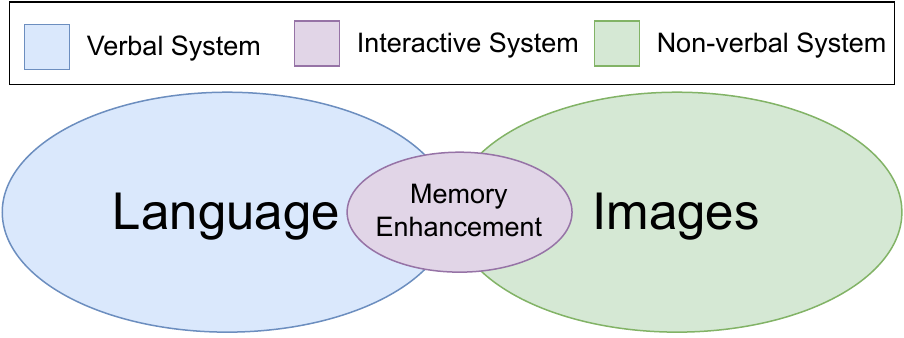}
    \caption{Dual Coding Theory}
    \label{fig:dual_coding}
\end{figure}

To reveal how multimedia assists in learning, it is critical to understand humans' information processing system and how this system formulates learning.
\citep{paivio1971imagery} believe that humans possess two cognitive subsystems: a verbal system to process linguistic information via media, such as language text, and an imaginal system to charge non-verbal information, such as image sensory. While the prior is good at sequential and symbolic information processing, the latter is more at holistic and spatial information processing. While the two systems are distinct, they also collaboratively interact to enhance memory and sense-making. For example, when students read the word force, they may simultaneously generate an image representing force or the scenarios in which they encountered force. Due to the difference between students’ prior knowledge and experience, the visualization in students’ minds can be distant, which needs diverse learning aids to facilitate meaningful learning (See Fig.~\ref{fig:dual_coding}).
In addition, \cite{paivio1971imagery} hypothesized that when students learn from multimedia with verbal and visual explanations of a natural phenomenon, they first build mental representations of verbal and visual systems, respectively, and construct referential connections between the two mental representations. The individual differences in the dual coding process result in different learning styles. \cite{clark1991dualcoding} suggest that Paivio's model pictured "the mental processes that underlie human behavior and experience. Paivio's theory was characterized as a Dual Coding Theory based on the features.

Aligning with the dual coding theory, \cite{mayer1994picture} further pointed out that building mental representations of external sensory stimulations and constructing referential connections between the visual and verbal representations happen in students' working memory. Given the constraints of humans' working memory, each channel (i.e., auditory and visual) has a limited capacity for processing information. Overloading any one channel can impede learning and information retention. Therefore, education should consider the intensity of information provided to learners and based on which to aid student learning. For example, when students read a text-based problem to figure out why the sun rises in the east and sets in the west, they must first transform the text into visual information, retrieving everyday experience from long-term memory. Meanwhile, they must retrieve scientific knowledge about circular motion and Newton's Laws in text and image \citep{zhai2022model}, which can be highly cognitively intensive. Therefore, appropriate strategies to assist problem-solving are essential to facilitate learning. In light of this,\cite{mayer1994picture} presents a list of principles for multimedia learning:

\begin{itemize}
   \item \textbf{Multimedia:} People learn better from words and pictures than from words alone. This principle supports using combined text, images, and audio for effective learning.
    \item \textbf{Contiguity:} People learn better when corresponding words and pictures are presented closely rather than separately. This means that text and images should be integrated rather than separated on different pages or areas of the screen.
    \item \textbf{Modality}: People learn better from graphics and narrations than from animation and on-screen text. This principle suggests that explaining visual material via audio narration is more effective than text that competes for visual attention.
    \item \textbf{Redundancy:} People learn better from graphics and narration than graphics, narration, and on-screen text. Excessive on-screen text can overload the visual channel.
    \item \textbf{Coherence:} People learn better when extraneous material is excluded rather than included. This principle advises against unnecessary content that does not directly support the learning objectives.
    \item \textbf{Personalization:} People learn better when words are presented in a conversational rather than a formal style and when the narration is in a friendly human voice rather than a machine voice.
    \item \textbf{Segmentation:} People learn better when a multimedia lesson is presented in user-paced segments rather than as a continuous unit. Breaking content into smaller chunks makes it easier to process.
    \item \textbf{Pre-training:} People learn better from a multimedia lesson when they know the names and characteristics of the main concepts. Pre-training in key concepts can improve understanding of the material.
\end{itemize}

Following this strand of research, many studies have examined the effect of multimedia learning compared to single-media learning. In a meta-analysis, the overall weighted mean effect size of multimedia learning compared to the control group was moderate to large, \textit{d} = 0.72 (95\% confidence interval 0.52–0.92)\citep{GINNS2005313}. One of the largest effect sizes was reported in \cite{moreno2002learning}, which tested the effect of multimedia learning in a computer-mediated botany class. According to the sub-experiment, the effect size was estimated to be \textit{d} = 2.11-2.52. The success of dual-coding or multimedia theory in learning can also be attributed to its affordance of reducing cognitive load, which could be higher when information is delivered only through the verbal channel \citep{sweller2011cognitive}.

In summary,  dual coding and multimedia theories emphasize audio and visual sensory perceptions, which could be used simultaneously. However, they have also empirically shown that the possibility of multimodal teaching and learning depends on students' information processing ability.

\subsection{VARK (Visual, Aural, Read/Write, and Kinesthetic) Multimodality Framework} \label{Gyeong-Geon}

While multimedia theory articulates the channels for information processing, researchers found that learning extends beyond media types and extends to communication modes. By comprehensively considering Visual, Audio, Reading/Writing, and Kinesthetic (VARK) modes, \cite{fleming1992not} proposed a more sophisticated VARK framework for multimodality learning. They grounded their assertions on cognitive science, such as neuro-linguistic programming, split-brain research, and left/right brain modalities \citep{fleming1992not}. Particularly, it is noteworthy that (1) \cite{fleming1992not} were deeply concerned with students' metacognition on their learning styles; (2) researchers differentiated 'visual' preferences for graphical and symbolic representation from 'Read/Write' for printed representation, which is more elaborated than the multimedia theory; and (3) researchers provided learning strategies that correspond to individual learners' preferred combinations of learning modalities. However, as stated by \cite{fleming1992not}, researchers primarily focused on the practical use of the inventory and did not provide construct validity or reliability of the instrument.
    
Posterior studies have revised and validated the VARK questionnaire \citep{leite2010attempted}, investigated the distribution of preferred learning modalities among students \citep{prithishkumar2014understanding}, and showed the effectiveness of multimodal teaching and learning based on the framework \citep{lee2019integrating}. The VARK multimodality is frequently referenced in educational technology and AIEd (Artificial Intelligence for Education) fields as well \citep{zhai2023ross}. For example, \cite{chen2015employing} showed that the VARK-inspired augmented-reality-embedded instruction effectively increased student achievement while dispersing imparities of individual learning styles. Also, \cite{lee2023development} developed the hands-free AI speaker system supporting hands-on science laboratory class, insisting that AI speaker complements the auditory learning channel to the previous laboratory course that utilized visual, reading/writing, and kinesthetic modes.

\subsection{Multimodal Assessment} \label{Yizhu}
In light of the multimodal features of learning, assessments are expected to meet the needs of using multimodal evidence to better reveal students' thinking in applying their knowledge to problem-solving.  Traditionally, assessment tasks focusing on factual knowledge are broadly used to assess students’ knowledge \citep{kuechler2010performance}. Such assessment tasks include multiple-choice questions, written explanations, essays, or short answers. The information gained from these assessments can support instructors in making instructional decisions, ultimately bolstering student learning \citep{pellegrino2006rethinking}. However, assessments using a single mode to represent information have limited capabilities to measure student learning \citep{smith2018multimodal} comprehensively. For example, the Next Generation Science Standards (NGSS) emphasize engaging students in scientific practices throughout science education \citep{national2014developing}. Accordingly, assessments should evaluate students' proficiency in employing scientific practices (e.g., conducting investigations and designing solutions). Multimodal assessments offer a promising avenue to gather evidence on how students engage in inquiry and engineering design \citep{fjortoft2020multimodal, Zhai2023eric}.
    
Due to the growing advancement and accessibility of information and communications technologies, assessments evolve to encompass more than just texts for posing questions, requiring students to incorporate various modes such as written responses, visual images, and interactive elements into responses. This shift in assessment strategies reflects the evolving landscape of educational technology, enabling assessments to align more closely with contemporary learning methodologies and encouraging multifaceted modes of student expression and comprehension. For example, the Programme for International Student Assessment (PISA) implemented by the Organisation for Economic Co-operation and Development (OECD) frequently utilizes a combination of written texts and illustrations to present natural or social phenomena, prompting students to synthesize information from various modes in their explanations. Moreover, advancements in educational technologies have facilitated the collection of multimodal responses to assessment tasks \citep{zhai@2020review}. Students' responses are not confined to textual answers but can take various formats, such as drawings, modeling, and running simulations, to convey their understanding. For example, \cite{zhai2022applying} asked students to create a drawn model and provide a written explanation of their model.  \cite{du2023multimodal} adopted a multimodal analysis of college students’ collaborative problem-solving in computer networks. Rather than requiring students to write answers, the task asks students to interact with a simulation program to check their IP addresses and then collaborate with their team members to build a network topology. This task requires multiple inputs with varying modalities.
    
There are several advantages of multimodal assessments. Multimodal assessments bolster construct validity by aligning more closely with the multifaceted nature of learning. By incorporating diverse modes such as written responses, visual representations, and interactive elements, these assessments more accurately measure what they intend to assess \citep{tan2020assessing}. This alignment ensures assessments effectively capture the breadth and depth of students' knowledge, skills, and understanding across various dimensions, thereby enhancing the credibility and accuracy of assessment outcomes. Moreover, multimodal assessments foster increased engagement among students. In addition, given students' preferences for information channels, multimodal assessments provide increased opportunities to express their thinking and engage in learning, facilitating educational equity and diversity \citep{zhai2023ross}. As technological advancements continue to reshape educational paradigms, the convergence of multimodal assessments and AGI holds immense potential for revolutionizing how learning is evaluated.

\section{Artificial General Intelligence}\label{Ehsan，SHUCHEN GUO}
As the field evolves toward AGI, multimodality has become a fundamental concept to meet this goal \citep{gemini2023}. By means of AGI, machines are expected to comprehend and incorporate diverse input and output modalities, including but not limited to visual, auditory, textual, graph, geospatial vector data, etc. \citep{radford2021clip,huang2023kosmos1,mai2023opportunities}. A multimodal AGI \citep{radford2021clip,huang2023kosmos1,peng2023kosmos2,openai2023gpt,balsebre2023cityfm,mai2023csp} is skilled at comprehending and interacting with various data types, unlike traditional AI systems that might specialize in a single modality (like text or images) \citep{ouyang2022instructgpt,Birhane2022gpt}. Thanks to this ability, it can recognize and understand the world from various aspects and react in ways that suit the interaction's circumstances. For example, a multimodal AGI such as GPT-4Vision \citep{openai2023gpt4} and KOSMOS-2 \citep{peng2023kosmos2} could read text, comprehend spoken language, and analyze visual data from a video simultaneously, combining these various inputs to create a thorough understanding. This wide range of sensory and processing capabilities is essential for developing a more human-like perception of reality and allowing AGI systems to function well in intricate, real-world settings containing various information sources. AGI's multimodality lies in processing various types of data and their seamless integration, emulating human cognition's capacity to combine sounds, images, and other sensory inputs to create a coherent experience. This approach does not replicate human-like cognitive abilities. Instead, it combines various sensory modalities to improve learning experiences \citep{legg2007collection}.

Replicating human intelligence and problem-solving abilities is central to the traditional definition of AGI \citep{wang2019defining}. However, in learning environments, the focus switches from AGI's human-like flexibility and autonomy to how AGI can interact with students \citep{zhai2021review}. Multimodal AGI systems perform exceptionally well in dynamic educational environments because they can adjust to different teaching and learning pedagogies and meet the individual needs of every student \citep{lake2017building, cao2023elucidating}. In contrast to single-modal AI, which is typically thought of as functioning autonomously, multimodal AGI in education works best when it collaborates with human educators, supporting and enhancing their pedagogical approaches.

Furthermore, multimodal AGI deviates from the singular and adaptive intelligence \citep{kahneman2011thinking}; instead, it integrates various specialized skills that cooperate to produce a learning environment that is more efficient and adaptable \citep{goertzel2014engineering}. AI programs that evaluate written assignments, decipher spoken inquiries, or graphically explain difficult ideas are examples of this \citep{wilson2023}. A wider variety of learners can access and enjoy education because each mode of interaction provides a unique point of entry for learning.

As AGI in education advances, increased attention is being paid to developing technologies that enhance and supplement the educational experience rather than developing systems that replicate human thinking processes \citep{wang2019defining}. This change recognizes AGI's special advantages and uses them to produce more dynamic, inclusive, and useful teaching resources. The result is a type of AGI that is better suited to satisfy the changing needs of educators and students alike, even though it may not be as closely aligned with the traditional goals of AGI \citep{legg2007collection}.

\subsection{Essential Components of Artificial General Intelligence}

AGI aims to imitate human intelligence in computers through a wide range of mechanisms and capabilities. The development of complex AGI systems depends on various cognitive processes, learning capacities, decision-making abilities, and adaptation strategies, which are the fundamental components of AGI \citep{goertzel2014engineering}.
% (See Fig.~\ref{fig:enter-agi_tree}).
% \begin{figure} [htp]
%     \centering
%     \includegraphics[width=0.75\linewidth]{figures/agi_tree.jpg}
%     \caption{AGI tree}
%     \label{fig:agi_tree}
% \end{figure}
\textbf{Constructing Cognitive Frameworks:} The creation of cognitive frameworks is crucial in AGI development. These all-encompassing systems incorporate cognitive processes like reasoning, learning, memory, and perception. These frameworks allow AGI systems to imitate human cognitive processes by modeling the structural and functional principles of the human mind \citep{langley2006cognitive}.

\textbf{Representation and Manipulation of Knowledge:} Capabilities to represent and manipulate knowledge are essential to AGI. This entails accumulating knowledge about the world, applying logic, making deductions, and updating it in light of fresh experiences \citep{davis2015commonsense}. Logical reasoning, probabilistic models, semantic networks \citep{speer2013conceptnet,ballatore2013osm}, and knowledge graphs \citep{auer2007dbpedia,vrandevcic2014wikidata,regalia2018gnisld,noy2019industry,janowicz2022know,qi2023evkg,shi2023chatgraph} are some of the techniques that AGI systems use for this purpose.

\textbf{Adaptive Learning Capabilities:} AGI needs to be flexible and experience-based. This entails employing machine learning methods like deep reinforcement and unsupervised learning to change behavior and enhance performance \citep{bommasani2021opportunities, goodfellow2016deep}.

\textbf{Strategic Planning and Decision-Making:} For AGI systems to accomplish tasks and solve challenging issues, they must be capable of advanced planning and decision-making. Plans are created using search algorithms, optimization, and game theory to optimize desired results \citep{norvig2016artificial,zhai2021advancing}.

\textbf{Language Processing and Interaction:} For AGI to process information and communicate with people efficiently, it must comprehend and produce human language. For this, natural language processing techniques such as discourse analysis, semantic analysis, and syntactic analysis are applied \citep{ouyang2022instructgpt,openai2022chatgpt, hirschberg2015advances, zhai2022applying}.

\textbf{Integrating Multimodal Data:} Text, image, video, and audio data are just a few sources AGI systems have to handle \citep{peng2023kosmos2,mai2023csp}. Effective decision-making in AGI requires extracting and synthesizing information from these multimodal inputs \citep{zhao2023brain}.

Together, these principles and methodologies underpin the development of AGI, aiming to create intelligent agents capable of functioning and learning in ways akin to human intelligence. 

\subsection{Multimodality in AGI vs. Artificial Narrow Intelligence}
AGI and artificial narrowing intelligence (ANI) represent two distinct paradigms in artificial intelligence. AGI refers to a form of intelligence that can understand, learn, and apply knowledge in a general, human-like manner, capable of performing any intellectual task that a human being can \citep{mclean2023risks}. In contrast, ANI, also known as "weak AI," is specialized in a specific task or field and lacks the general cognitive abilities of AGI \citep{korteling2021human}.

Multimodality, the ability to process and integrate multiple data types (visual, auditory, and textual), is a key area where AGI and ANI significantly differ. While early pioneers of AGI systems such as GPT-3 and ChatGPT can only handle one modality, many recent multimodal AGI systems such as KOSMOS-2 and GPT-4Vision were dedicated to handling various data modalities, mimicking the human brain's ability to process diverse sensory inputs \citep{legg2007collection,xiao2023instruction,openai2023gpt4,wang2023review}. This capability allows AGI systems to perceive the environment holistically, leading to more nuanced understanding and decision-making \citep{kahneman2011thinking}.

Moreover, leveraging pretraining on extensive multimodal datasets and advanced reasoning capabilities of Large Language Models (LLMs), the newly released AGI models exhibit emergent abilities in processing previously unseen data \citep{yin2023survey}, aligning closely with human zero-shot and few-shot learning capabilities \citep{zhao2023brain}. The multimodality of AGI systems significantly enhances their robustness in interacting with the environment and following human instructions that may encompass text, voice, gestures, and even brain signals \citep{chen2023minigpt,wu2023next,han2023imagebind,liu2023visual,awadalla2023openflamingo,wei2023chat2brain,zhao2023brain,cai2023multimodal,zhou2023fine}.

The multimodal capabilities of AGI confer a significant advantage in terms of performance and adaptability. AGI systems can operate in various environments and tasks, adapting to new challenges with a human-like understanding \citep{zhai2021review}. This flexibility starkly contrasts the rigid, task-specific functionality of ANI systems, which are optimized for specific applications but cannot generalize their knowledge to new contexts \citep{lake2017building}.

Furthermore, AGI’s multimodal integration enables it to make more informed and accurate decisions. By synthesizing information from various sources, AGI can assess situations with a level of depth and context that ANI cannot achieve \citep{goertzel2014engineering}. This makes AGI particularly valuable in dynamic and unpredictable environments, where the ability to adapt and respond to new information quickly is crucial.

In the educational domain, AGI's multimodal capabilities are particularly relevant \citep{milano2023large,extance2023chatgpt,kumar2023math}. They can be applied in various educational scenarios, such as language learning through listening and speaking exercises, supporting the creation of mind maps and flowcharts for writing and reflection, and modifying, clarifying, and contextualizing instructions for the gamification of learning activities. Additionally, AGI can provide multimodal samples for students to observe and analyze genre features, such as infographics in multimodal writing. For instructors, AGI is an invaluable tool in syllabus design, activity brainstorming, and assignment creation. It can transform text-based syllabi into multimodal formats with timelines, flowcharts, and maps, making it easier for visual learners to digest. Furthermore, AGI models can be a robust teaching assistant by providing multimodal feedback beyond traditional written feedback, like audio feedback. AGI's extensive applications and influence in education are further explored in the following Sections.

To sum up, the multimodal nature of AGI sets it apart from ANI, providing it with a broader range of capabilities and a more flexible approach to problem-solving. While ANI excels in specific tasks, AGI's general-purpose intelligence and ability to integrate multiple data types make it more suited for complex, real-world applications that require a nuanced understanding of diverse inputs.

\section {Text Analysis and Generation in Education} \label{Arne Bewersdorff}
\subsection{Definition of Text and Its Role in Education}
The text combines characters and symbols to transfer, store, synthesize, and analyze information \citep{Thompson2011}. Text is a central modality in our information-driven world. Since most of the information we consume and produce is text-based, its role in education becomes indispensable. In education, `text' refers to any form of written content used as a medium for imparting education. This includes many materials such as textbooks, scholarly articles, digital content, interactive media, and various other written formats \citep{Russell2009Context}. Written text is not only a fundamental part of education but also serves as one primary medium through which students acquire knowledge and develop critical thinking skills \citep{Allison1998,Bates.2015}. Therefore, AGI, capable of generating and analyzing text, is transformative in education.

\subsection{Technical Background}
For text analysis and generation, AI systems based on LLMs are becoming extremely popular \citep{li2022pretrained, dathathri2019plug}. Before LLMs, Recurrent Neural Networks (RNNs) \citep{tarwani2017survey} that process data sequences by maintaining a memory of previous inputs were popular for natural language processing (NLP) tasks. However, RNN faces limitations in handling long-term dependencies. This led to the attention mechanism's development, allowing models to focus on different parts of the input sequence for a better understanding of long-range dependencies. The Transformer architecture, introduced in 2017, marked another significant leap forward \citep{vaswani2017attention}. It utilizes self-attention to weigh the importance of different parts of the input data and consists of alternating layers of self-attention and fully connected neural networks, proving more effective than RNNs for many NLP tasks. BERT (Bidirectional Encoder Representations from Transformers, \citep{devlin2018bert}, a model using the Transformer's encoder mechanism, further revolutionized NLP by understanding the context of words in sentences bidirectionally. LLMs like OpenAI’s GPT-4 \citep{OpenAI2023GPT4TR} have scaled this concept, showing that the effectiveness often increases with model and training corpus size.

\subsection{Generative Large Language Models}
The Generative Pre-trained Transformer (GPT) series, comprising GPT \citep{radford2018improving}, GPT-2 \citep{radford2019language}, GPT-3 \citep{brown2020language}, ChatGPT/GPT-3.5 \citep{openaiIntroducingChatGPT} and GPT-4 \citep{openai2023gpt4}, are decoder-only generative language models, each building upon the advancements of its predecessor. These models, based on the transformer architecture \citep{vaswani2017attention}, primarily focus on text generation tasks. 

In a standard transformer, there are two main components: an encoder and a decoder. The encoder processes the input data and the decoder generates the output. However, in the GPT models, only the decoder part is used.

Each layer of the decoder includes two main sub-layers. The first is a multi-head self-attention mechanism, and the second is a position-wise fully connected feed-forward network. The multi-head self-attention mechanism allows the model to focus on different parts of the input sequence, which is crucial for understanding the context and relationships between words in a sequence.

The attention mechanism in transformers, including GPT models, is based on queries (Q), keys (K), and values (V). The attention scores are computed by taking the dot product of the query with all keys and then applying a softmax function to obtain the weights on the values. In a multi-head attention setup, this process is done in parallel with different, learnable linear transformations of Q, K, and V, allowing the model to jointly attend to information from different representation subspaces at different positions.

\begin{equation}
Attention(Q, K, V)=softmax(\frac{QK}{\sqrt{d_k}})V
\end{equation}
where $d_k$ denotes the dimension of \textbf{K}.

Layer normalization \citep{ba2016layer} and residual connections \citep{he2016deep} are also employed in each sub-layer, enhancing training stability and allowing for deeper models. The output of each sub-layer is LayerNorm(x + Sublayer(x)), where Sublayer(x) is the function implemented by the sub-layer itself. This architecture enables generative language models models to effectively capture complex patterns and dependencies in data, contributing to their impressive performance in generating human-like text. GPT-2 and GPT-3, for instance, demonstrate the scalability of this architecture. GPT-2, with 1.5 billion parameters, was a leap from GPT in terms of size and training data. GPT-3, significantly larger with 175 billion parameters, further pushed the limits of text generation capabilities, handling complex tasks like translation, question-answering, and even code generation with unprecedented proficiency. Non-GPT generative language models, such as PaLM-2 \citep{anil2023palm} and the LLaMA series \citep{touvron2023llama}, have also experienced similar success.

The release of ChatGPT in November 2022, a conversation-focused model that follows human instructions, further underscored the feasibility of AGI in practical applications \citep{liu2023summary}. This development has had a wide-ranging impact across various sectors, including journalism \citep{liu2023transformation}, education \citep{RN3362,liu2023context}, healthcare \citep{li2023artificial,liu3surviving,holmes2023evaluating}, industry \citep{dou2023towards}, agriculture \citep{rezayi2023exploring}, law \citep{bubeck2023sparks}, gaming \citep{bubeck2023sparks}, and finance \citep{wu2023bloomberggpt}, catalyzing a popular wave in AI \citep{liu2023summary,liu2023radiology,liu2023evaluating}.

GPT-4, the most recent iteration, stands as the most powerful LLM to date. Its enhanced capabilities and performance have brought the impact and potential of AI into sharper focus, making the public more aware of the transformative power of these technologies  \citep{openai2023gpt4}. This progression in the GPT series not only reflects the rapid advancements in AI and NLP but also underscores the increasing importance of these technologies in various aspects of modern life \citep{zhao2023brain,liu2023summary,liu2023transformation,liu2023pharmacygpt,liu2023radonc}.

\subsection{Potentials of LLMs in Text Analysis and Generation in Education}
LLMs have been broadly used in text analysis and generation in education from the teachers’ and learners' perspectives.

\subsubsection{Teachers’ perspective}
\textit{For Assessment and Feedback.} Assessment has been, despite some critics \citep{Bennett2010}, identified as one of the most significant influencing factors for effective learning \citep{Hattie2009}, especially forms of (computer-based) `rapid formative assessment' is highly effective \citep{Yeh2010}. AGI-driven systems could potentially transform assessment practices and help teachers build on this feedback \citep{SwieckiEtAl2022,zhai2021practices}.

Since LLM-based AGI systems such as InstructGPT \citep{ouyang2022instructgpt}, OPT \citep{zhang2022opt}, and GPT-3 \citep{openai2023gpt} are trained to generate high-quality texts, they can assist users in their writing process \citep{Yuan2022Wordcraft}. More specifically, LLM systems such as ChatGPT can help improve writing skills from a very young age to professional writing \citep{zhai2022gpt}. During their academic years, students are learning various types of essays. However, teachers cannot provide detailed feedback for each student’s work in the traditional classroom setting due to time constraints and heavy workloads. AGI systems like the PEER system \citep{sessler2023} can assist the teacher with assessment and providing feedback on student’s essays. This stands in contrast to previous work, where the focus was often on merely grading the essay rather than offering comprehensive feedback \citep{RameshSanampudi2022}.

Identifying logical errors in complex, incomplete, or even contradictory and overall heterogeneous data like students’ written data like experimentation protocols is challenging. Current evaluation methods are mostly pen-and-paper based, time-consuming, and complex \citep{Baur2015}. LLMs can automatically identify student errors in protocols as reliable and valid as an expert \citep{Bewersdorff2023} and provide individualized feedback \citep{sessler2024}, streamlining teacher assessments.

\textit{For Curriculum development and content creation. }Finding or developing suitable learning opportunities for students is the foundation of every learning process. LLMs can not only create exercises with sample solutions but also complementary guiding materials and explanations (computer science: \citep{Sarsa2022})

Creating assessments is essential for school teachers, particularly for creating formative and summative assessments. High-quality tasks enable teachers to diagnose student conceptual understanding and difficulties, monitor progress, and evaluate the efficacy of pedagogical methods. AGI-driven applications can help teachers create tasks (science education: \citep{Kuchemann2023}) or multiple-choice questions (language learning: \citep{RainaGales2022}) for students’ assessment.

\subsubsection{Learners' perspective}
\textit{To stimulate affective variables like curiosity and motivation}. One of the essential prerequisites for learning is developing and maintaining learning motivation \citep{ZimmermannSchunk} and underlying curiosity \citep{Silvia2012}. \cite{AbdelghaniEtAl2023} explored using GPT-3 to stimulate children's curiosity and improve their ability to ask questions. They developed a system that automatically generates prompts designed to encourage children to ask more and more profound questions. The findings suggest that LLMs can be effective in promoting curiosity-driven learning.

AGI-driven systems can help improve students’ motivation by providing an interactive and engaging learning experience. \citep{SilitongaEtAl2023} demonstrated the role of LLM-based learning in boosting students' motivation for English writing. They observed that text-based AI chatbots offer more prosperous and nuanced feedback than teachers alone, motivating students to engage more deeply with their learning.

\textit{To improve peer assessment.} Peer assessment is a process by which students give feedback on another student’s work based on a rubric provided by the instructor \citep{SadlerGood2006}. Results indicate that peer assessment is more effective than teacher assessment and remarkably robust across a wide range of contexts \citep{DoubleMcGraneHopfenbeck2020}. AGI-driven systems can comprehensively and automatically evaluate the quality of peer review comments to improve peer assessment \citep{Jia2021}.

\textit{To foster personal and adaptive learning.} Adaptive learning can significantly improve students' learning process (e.g., \citep{Elfeky2018}). A key advantage of text-based LLMs over static sources of information, such as textbook representations, is the adaptive nature of these systems. Whereas in the past such support had to be provided by the learning material, it is now possible for learners to generate such support adaptively.

\subsection{Outlook}
The shift from simple digital tools for text teaching to advanced AGI-driven systems capable of complex text-related tasks, such as evaluating student writing and generating educational content, highlights the growing importance of text-centric AGI in education.

Incorporating AGI-driven systems in education marks a significant transformation in the field. These systems, primarily focused on text analysis and generation, offer comprehensive benefits for educators and students. They enhance various aspects of learning, including providing detailed text-based feedback, assisting in developing text-rich curricula, and boosting student motivation through interactive textual engagement.

Looking forward, the potential of text-based AGI-driven systems in education is vast. The future points towards personalized text-based learning assistants, offering personalized support by analyzing and responding to students' textual inputs. These systems could be integrated with other modalities like audio, images, and video, making these systems more engaging and interactive. This progression emphasizes the evolution of AGI from a tool supplementing text-based education to a central and multi-purpose agent of the educational ecosystem.

\section {Graphics Analysis and Generation in Education}\label{Gyeong-Geon-GengchenMai-Yizhu} 

\subsection{Image Classification}
Image classification is a fundamental task in computer vision that teaches machines to categorize images into one of several predefined classes. Traditionally, a dual-stage approach was used to solve the classification problem \citep{rawat2017deep}. Handcrafted features were first extracted from images using feature descriptors (e.g., edge directions, texture patterns), and these features served as inputs to a trainable classifier. Feature extraction can be automated by the histogram of oriented gradients (HOG), scale-invariant feature transformation (SIFT), and local binary patterns (LBP) approaches. They differ in their primary focus and methodology. HOG focuses on edge and shape information, SIFT specializes in key point detection invariant to transformations, and LBP captures local texture patterns within images \citep{ahmed2019content}. Machine learning algorithms such as support vector machines, random forests, and K-nearest neighbors, are conventionally utilized to address image classification tasks. However, the accuracy of the classification process heavily relied upon the feature extraction stage \citep{xiao2015application}. This restricts the adaptability of these algorithms to complex datasets with variations in scale, orientation, and lighting.

After developing the convolutional neural network (CNN), image processing technology faced a huge turn. CNN receives 2-D pixel data, usually comprising three color layers (Red, Green, and Blue), as input and returns the convolutional filter layer whose parameters are updated according to the training dataset and loss function. CNN models have shown remarkable performance in learning hierarchical features from images \citep{rawat2017deep}. CNN models typically consist of convolutional layers, pooling layers, and fully connected layers \citep{rawat2017deep}. They automatically extract relevant features from images through convolutional and pooling operations, reducing the need for handcrafted feature extraction. CNN models utilize backpropagation and optimization algorithms to learn model parameters, making them adaptable and capable of learning complex patterns \citep{chen2021review}. Nowadays, typical image processing pipelines use CNN layers as modules in the larger model. 

Image classification models have automatically scored student-generated images \citep{zhai2022applying,lee2023automated,wang2024modeling}. \cite{von2023scoring} adopted artificial neural networks to classify graphical responses from a TIMSS 2019 item. They found that CNNs outperform feed-forward neural networks in classification accuracy comparable to human raters. \cite{zhai2022applying} used ResNet-50 V2 CNN to develop scoring models for students' free-drawn modeling tasks. \cite{lee2023automated} developed a method employing instructional notes and rubrics to prompt GPT-4V to score students' drawn models for science phenomena. 

\subsection{Object Detection} \label{Gyeong-Geon}
Object detection is one of the most fundamental tasks in computer vision, which "deals with detecting instances of visual objects of a certain class (such as humans, animals, or cars)", solving the problem of "\textit{what objects are where}?" in the given image (p. 257; italic as in original) \citep{Zou2023OD}. As with other machine learning fields, object detection has also developed with current deep neural network-based algorithms - particularly CNN - since the 2010s \citep{Zhao2019OD}. Deep neural network-based object detectors have developed in two strands: (1) a Two-stage detector exemplified as RCNN (Regions with CNN features)\citep{Girshick_2014_CVPR} that first searches object proposals in the image and then applies another layer that extracts feature, and (2) One-stage detector exemplified as YOLO (You Only Look Once)\citep{Redmon_2016_CVPR} that prioritizes processing speed using a single neural network to the image. Recently, one-stage detectors supplemented with high accuracy are becoming more prominent \citep{Zou2023OD}. For example, YOLO is one of the first algorithms that enabled real-time object detection even in mobile smartphones and its accuracy is constantly increasing. This real-time object detection is connected to object tracking since video can be considered as the flow of images.
    
The domains object detection has been applied can be identified by looking into the benchmark datasets such as VOC, ILSVRC, MS-COCO, Objects365, and OID \citep{Zou2023OD}. Images in these datasets are annotated with real-life objects such as "person", "cat", "bicycle", "airplane", "apple", "microwave", "pencil box", "chair", etc. However, although they mostly emphasize real-world situations in object detection, there is almost no benchmark image data for object detection, particularly for educational settings. Thus, models are specialized for educational purposes.
    
Studies have suggested adopting object detection for educational studies. Many technically demonstrated that object detection can be used to recognize students' face, hand raising, sleeping, standing, attention, etc. in classrooms \citep{Xu2019classroom, teeparthi2021longterm, su2021learning, shao2018multiobject, musambo2018student, wu2021recognition, xu2023analyzing}. However, most of them have been exploratory, and only a few studies have reported their performance metrics other than detection frequency \citep{shao2018multiobject}. Notably, \cite{gamifying-math-education} developed the phygital (physics + digital) curriculum-inspired board game for kids aged 5-8. In the game, students could manipulate wooden blocks to assemble various polygons, and then the camera detected and predicted student-assembled polygons almost perfectly (F1 > .99). Also, \cite{Lee_2019_KASE} suggested that YOLO could be used for detecting experimental apparatus and equipment in laboratory education setting to support safe laboratory learning activities.
    
To sum up, object detection techniques for 2D images have not been applied in various ways for education. However, techniques developed for object detection work as a basis for visual question answering and video analysis, which are very promising, as explained below.    

\subsection{Visual Question Answering} \label{Gyeong-Geon}

As one type of question-answering task \citep{bouziane2015question,mishra2016survey,mai2021geographic}, Visual Question Answering (VQA) was introduced by \cite{antol2015vqa}, as a new problem that advanced multimodal AI should be incorporated to solve and could be evaluated by it. \cite{antol2015vqa} defined that "VQA system takes as input an image and a free-form, open-ended, natural-language question about the image and produces a natural-language answer as the output" (p. 2425). For example, an AI model is given an image and a user query, such as 'Does this man have children?', 'what kind of cheese is topped on this pizza?' and 'How many glasses are on the table?'. Then, the model could answer like 'yes', 'mozzarella', and '6' \citep{antol2015vqa}. To solve these open-ended VQA tasks, the multimodal AI should be able to make text-based Q\&A and describe visual content, which requires the integration of computer vision, natural language processing, and knowledge representation and reasoning. Therefore, VQA can be considered a more sophisticated task than object detection and image captioning \citep{antol2015vqa, teney2017visual}. \cite{antol2015vqa} trained their milestone model with more than 200,000 images from the MS-COCO dataset, combined with more than 750,000 questions and almost 10,000,000 answers written in natural language. Their best model (LSTM + Question + Image) showed an average accuracy of 54.06\% throughout open-ended and multiple-choice questions. Afterward, VQA techniques have been developed on the strength of attention mechanisms, transformer, pre-trained language models, memory-augmented neural networks, and run-time retrieval of additional information \citep{teney2017visual, kafle2017visual, wu2017visual, sants2023visual}. For the benchmark data extended from \cite{antol2015vqa}, models have shown more than .70 of accuracy in 2016 \citep{fukui2016multimodal} and .8283 in 2022 \citep{bao2022vlmo}. Also, other benchmark datasets such as DAQUAR, COCO-QA, Visual Genome, Visual7W, and CLEVR have been suggested, and many models have been competing for the state-of-the-art, showing more than .70 accuracy.

Notably, one of the motivations to develop VQA technology was the consideration for visually impaired learners \citep{bigham2010vizwiz, antol2015vqa}. \cite{bigham2010vizwiz} suggested that it would be very helpful to visually impaired learners if we could provide them with real-time natural language-based explanations of the scene they face. They connected visually impaired learners with remote human workers through quikTurkit so that the people who need help can ask questions about their situation with photos and voices taken by their iPhones. For example, if the visually impaired person takes a photo of three cans and asks, "Which can is the corn?" remote human workers could return an answer like "the rightmost one." After experiencing human workers' remote answers, 11 visually impaired iPhone users responded that VizWiz is accurate and useful. The case of \cite{bigham2010vizwiz} exemplifies that VQA could serve the promise of multimodal learning. Unfortunately, even after developing VQA technologies using machine learning (not human workers), only several studies tried VQA for educational purposes.

\cite{aishwarya2022stacked} technically showed that Re-attention and BERT-based model could achieve 67.15\% of accuracy on VQA for textbook illustration. Similarly, \cite{gupta2023eduv} suggested that students could learn themselves using the VQA system with textbook illustrations - for example, an elementary student could ask a model "what kind of animal classes can be seen in this image?" and the model returns "mammal". \cite{suresh2018gamification} used VQA to the gamification of early learners' (3-4 years old) word learning by developing a system that asks a user to answer whether a presented image is an 'animal' or a 'sport' (Level 1), or which animal or which sport (Level 2). The novelty of this study is that the VQA model allows students to make diverse answers to the given image, as far as it can detect that object in it. \cite{sophia2021edubot} compared chatbot and VQAbot to enhance student learning related to weather and atmosphere - however, the VQA model has shown relatively low accuracy. Recently, \cite{lin2023research} suggested that VQA models with higher performance could benefit college students learning English, Mathematics, and Chinese. However, the details of educational measurement are not provided.

The number of VQA-based educational studies could have been limited by the technical barrier. Recently, OpenAI released GPT-4V and its API \citep{openai2023chatgpt}. Thanks to its powerful VQA functions and friendly chat-like interface, users with no programming experience could use VQA. While almost no research used GPT-4V for education, \cite{lee2023nerif} developed a Notation-Enhanced Rubric Instruction for Few-shot Learning (NERIF) method for automatically scoring student-generated images. They showed that GPT-4V successfully retrieves problem context and human evaluators' scoring examples in the image and extracts and explains student-drawn models for natural phenomena related to chemical reactions. As a result, they reported that GPT-4V's average scoring accuracy was 51\% for the trinomial image classification task according to students' learning progression - i.e., one of 'Proficient', 'Developing', and 'Beginning'.

To sum up, VQA is one of the most advanced multimodal AI technologies in the current era, incorporating NLP and CV. However, despite its potential, VQA technologies are still under development, and a small number of specific domains, such as medical sciences and remote sensing, have adopted VQA. Since the release of GPT-4V made utilizing it much easier, future studies on VQA for education are strongly recommended.

\subsection{Graphics Generation} 
\label{GengchenMai}

Generative models of images \citep{van2016pixelcnn,kingma2013vae,kobyzev2020normalizingflow,creswell2018gan,mildenhall2021nerf,he2021spatial,mai2023ssif,saharia2022imagen,rombach2022stablediffusion,ramesh2021dalle} become increasingly popular, especially after the development of generative adversarial networks (GAN) \citep{creswell2018gan}. The recent success of various large generative AI models such as ChatGPT, GPT-3, GPT-4Vision, DALL$\cdot$E \citep{ramesh2021dalle}, Imagen \citep{saharia2022imagen}, Stable Diffusion \citep{rombach2022stablediffusion} also foster the wide acceptance of generative AI of images to the general public. In contrast to the idea of discriminative models which mainly focus on modeling the conditional probability $P(\mathbf{y} | \mathbf{x})$ ($\mathbf{x}$ is the input data such as images, texts and $\mathbf{y}$ is the prediction target such as image labels), generative models aim at directly modeling the data distribution $P(\mathbf{x})$ for the unconditional generation or the joint probability distribution $P(\mathbf{x}, \mathbf{y})$ for the conditional generation. According to the fact whether a generative model can directly model the data distribution $P(\mathbf{x})$ (or $P(\mathbf{x}, \mathbf{y})$), we can in general classify generative models into two groups: likelihood-based generative models,  likelihood-free generative models, and score-based generative models. 

Likelihood-based generative models include many well-known generative model classes such as Autoregression models \citep{van2016pixelcnn}, Variational Autoencoders (VAEs) \citep{kingma2013vae}, Normalizing Flows \citep{kobyzev2020normalizingflow}, Neural Radiance Field (NeRF) \citep{mildenhall2021nerf,he2021spatial,mai2023ssif}, and Energy-based Models \citep{zhang2018overview}. All of these models directly model the data likelihood $P(\mathbf{x})$ (or $P(\mathbf{x}, \mathbf{y})$) so that it is very easy and efficient to do data likelihood evaluation and new data sampling (e.g., generating new images based on a pre-trained generative model). However, the disadvantage is that the neural network model architectures that can be used for these generative models are restricted. For example, Normalizing flow models requires invertible neural network architecture \citep{kobyzev2020normalizingflow}. Autoregressive models such as PixelCNN \citep{van2016pixelcnn} and GPT-3 \citep{openai2023gpt} require an order in the data need to be generated, e.g., $\mathbf{x} = [x_1, x_2, .., x_n]$ and a neural network model can handle this sequential dependency so that the data probability can be decomposed into a multiplication of a set of conditional probabilities, e.g., $P(\mathbf{x}) = p_{\theta}(x_1, x_2, .., x_n) = \prod_{i=1}^{n} p_{\theta}(x_i|x_{<i})$. Most large language models (LLMs) such as InstructGPT, GPT-3, OPT, and LLaMA-2 \citep{touvron2023llama} are autoregressive models by using Transformer \citep{vaswani2017attention} as the underlining backbone. These models are potentially very useful in education contexts. From instructors' perspective, we can use LLMs to grade text-based homework automatically. From the students' perspective, LLMs can be used to polish written essays and research papers and help in self-taught learning. While graphics generation can also be done by autoregressive models, VAEs, or normalizing flows, these models do not usually achieve the best generative performance.

As the name implies, likelihood-free generative models, or so-called implicit generative models \citep{mohamed2016learning}, do not explicitly model the data likelihood. Instead, the probability distribution of data is implicitly represented by a model of its sampling process. Examples are GANs \citep{creswell2018gan}. GANs use two intertwined neural networks -- the generator and the discriminator. The generator generates new data samples (e.g., images, texts) given a random noise or some conditions such as class labels. The discriminator aims to distinguish synthetic data samples generated by the generator from the real data samples. Two components are trained adversarially within a minimax game framework. Compared with the previous groups, GANs have a very flexible choice of model architectures and can do data sampling very efficiently. However, data likelihood evaluation is intractable and due to the adversarial training, the training process of GANs are usually unstable and usually leads to mode collapse issues. Before the wide acceptance of score-based models, GANs were the dominant models in graphic generation. In the education context, GANs have been shown to be effective to generate high-quality synthetic data samples which can improve data quality and expend dataset sizes in education research \citep{bethencourt2023use}.

Score-based generative models (SBMs) \citep{song2020score,saharia2022imagen,ramesh2021dalle,rombach2022stablediffusion,song2023consistency} are another set of generative models. Instead of directly estimating the data likelihood $P(\mathbf{x})$, they estimate the score function $\bigtriangledown_{\mathbf{x}}log p_{\theta}(\mathbf{x})$ - the gradient of the log of the data probability density function with respect to the data. Due to this property, SBMs are also called score-based generative models or diffusion models. The advantages of SBMs are four folds: 1) unlike likelihood-based generative models, score-based generative models can use very flexible model architecture; 2) unlike GANs, the model training is very stable; 3) SBMs can generate relatively high-quality samples; and 4) SBMs can be easily combined with other models to do zero-shot generations on tasks that they have not been trained on. For example, an unconditional image diffusion model can be combined with a pre-trained image classifier to generate a conditional image (e.g., generating an image by giving "dog" as the class condition). One unique advantage of SBMs compared with other generative models is that SBMs can be easily used for zero-shot image editing such as image inpainting \citep{song2020ddim,kawar2022ddrm}, image colorization \citep{kawar2022ddrm}, image super-resolution \citep{saharia2022image,song2023consistency}, image denoising \citep{song2020ddim}, stroke-guided image generation \citep{meng2021sdedit,song2023consistency}. However, one big drawback of SBMs is that the data sampling speed is very slow and several models such as DDIM \citep{song2020ddim} and Consistency Models \citep{song2023consistency} are proposed to speed up the data sampling process. Many popular image generative models are energy-based, such as Imagen, DALL$\cdot$E-2, and Stable Diffusion. The most well-known task these models can do is text prompt-based image generation. In education, these models can be used to generate synthetic illustration figures given the text inputs. Moreover, in the children's education domain, SBMs have been successfully used for generating children's storybooks given one plain text description \citep{jeong2023zero}. 

\section {Video and Audio Analysis and Generation in Education}\label{Mathew，SHUCHEN GUO}
Video analysis techniques can be applied in various educational settings, from K-12 to higher education and professional development \citep{gupta2023comprehensive}. On the other hand, audio in education covers spoken words and music as a tool for teaching and learning\citep{reddy2023audio}. This encompasses a wide range of formats and technologies, from traditional methods like lectures and radio broadcasts to modern digital audio formats like podcasts, audiobooks, and educational apps. 

The field of AI in education is currently experiencing a significant shift from ANI to the aspirational goal of AGI \citep{stangl2023potential}. However, the complete potential of AGI in education remains largely unrealized. Nevertheless, the advancements in ANI, especially in the areas of audio and video analysis, serve as a foundational platform for the eventual emergence of AGI \citep{anantrasirichai2022artificial, singh2019artificial}. Presently, ANI technologies are predominant in these areas: 

\subsection{Audio Analysis}
Audio analysis is widely supported by artificial intelligence technology such as speech, voice, music and environmental sound recognition. Several studies used audio analysis to analyze the classroom climate and discourse, etc.

For example, the study of \cite{james2019automated} indicated the potential of audio recordings for automatically predicting classroom climate by developing a system to detect speakers and social behavior and infer classroom climate from non-verbal features. \cite{canovas2022analysis} developed an automated classroom audio analysis system that can distinguish the speakers' identities (e.g., teacher, student) when there are multiple speakers simultaneously. Based on the audio information, they derive non-verbal features that can be used to describe patterns and classification of the different activities and teaching methods. Likewise, in their respective studies, \cite{cosbey2019deep} and \cite{benedetto2023abstractive} addressed two different yet crucial aspects of technological integration in educational settings. \cite{cosbey2019deep} introduced a set of deep learning classifiers designed for automatic activity annotation in classrooms. The deep learning classifiers analyze classroom recordings and categorize frames into "single-voice," "multi-voice," "no-voice," or "other," thereby quantifying different types of classroom activities. Their methodology proves to be significantly reliable, with a low frame error rate, demonstrating the potential of AI in enhancing the understanding of educational dynamics. Again, \citep{reddy2023audio}'s study on "Audio Classifier for Endangered Language Analysis and Education" examines two crucial aspects of their application: the audio classifier and language learning. The audio classifier is particularly adept at automatically identifying vowels and consonants in Blackfoot audio files, facilitating user access to these linguistic elements. Simultaneously, the language learning component of the application allows users to examine the pitch patterns of these instances visually. It is a useful tool for both research and education in the field of preserving endangered languages. Meanwhile, \cite{lee2023development} developed the AI-speaker system that recognizes students' spoken languages in the chemistry laboratory classroom and helps them calculate the number of needed reagents, recall experimental procedures, and determine where to dispose of liquid wastes.

\subsection{Audio Generation}

Moreover, on ANI, Wang et al. \citep{wang2024modeling} made a significant breakthrough with their innovative Transferable Audio-Visual Text Generation (TAVT) framework. Their framework aimed to understand and translate multi-modality contents into texts, addressing the challenge of integrating multiple forms of data, such as visual and auditory inputs. The TAVT framework comprises two key components: the Audio-Visual Meta-Mapper (AVMM) and Dual Counterfactual Contrastive Learning (DCCL). The AVMM establishes a universal auditory semantic space, aligning domain-invariant low-level concepts with visual representations, thus enhancing the audio-visual correlation. Their study significantly advances audio-visual text generation, offering a more integrated and accurate approach to processing and translating multi-modal data.

\subsection{Video Analysis}
Video can capture the complexities inherent in teaching and learning. With its feature of permanent record and allowing for detailed examinations of teaching and learning from multiple and different perspectives, there has been a notable increase in the use of video as a tool for analyzing classroom dynamics \citep{fitzgerald2013through}\citep{lv2021artificial}. And researchers start to apply AI, in video analysis due to the complexity of the task. For example, many studies have applied human activity recognition, face recognition, and emotion analysis to video analysis. These AI technologies help to track and examine classroom teaching and learning processes to provide useful information for teaching adjustment \citep{shenoy2022study}. 

Using computer vision techniques, \cite{van2020student} captured video of classroom situations, pre-processed it, and automatically classified the students into different emotion categories. \cite{sahla2016classroom} suggested a system for assessing classroom teaching by analyzing and classifying student mood. \citep{lee2022combining} combined deep learning and computer vision in an automated STEM activity behavior recognition system to aid understanding students’ learning processes in STEM activities.\citep{ramakrishnan2021toward} present a multi-modal automatic classroom observation system to analyze videos of school classrooms for the climate, based on integrating information including audio features, the faces of teachers and students, and the pixels of each image frame.

\cite{meline2023examining} conducted a meta-analysis to evaluate the effectiveness of video analysis in improving teacher outcomes. The results underscored the potential of video analysis as a promising practice, particularly in enhancing praise and implementation strategies in teaching. The study suggests that with methodological refinements and larger sample sizes, video analysis could become an evidence-based practice for educator development.

On a similar note, \cite{kubsch2017using} discussed how smartphone thermal cameras could address misconceptions about energy with simple experiments, showcasing the pedagogical utility of readily accessible technology. Meanwhile,\citep{dolo2018thermal} highlighted that while IR cameras can be powerful tools for inquiry-based learning in thermal science, their effectiveness is contingent upon students' pre-existing conceptual understanding. This nuanced perspective points to the need for foundational knowledge to fully leverage the potential of such innovative educational technologies.

\subsection{Video Generation}
In their influential work, \citep{wu2023tune} advanced in the field of AI-driven image and video generation. The models from their study are adept at generating still images that accurately represent verb terms, showcasing a remarkable ability to maintain content consistency even when generating multiple images concurrently. Building on this, the team introduced 'Tune-A-Video', a novel approach that incorporates a tailored spatio-temporal attention mechanism and an efficient one-shot tuning strategy to learn and replicate continuous motion. This breakthrough in text-to-video generation marks a substantial leap forward in the realm of AI-based media creation, offering new possibilities in how motion and action are represented and generated from textual descriptions.

On the other hand, \citep{benedetto2023abstractive} focus on optimizing lecture content summarization in their study which overcomes the limitations of traditional extractive models. This method is capable of generating more readable and coherent summaries of lecture content, proving to be a valuable tool in enhancing the accessibility and efficiency of learning materials for students. The use of abstractive summarization techniques reflects a significant advancement in educational technology, offering a promising avenue for future research and application in the field of e-learning and digital education. 

 \subsection{Audio-Video Generation}
 Ruan et al. (2023) introduced an innovative approach to audio-video synthesis with their Multi-Modal Diffusion model (MM-Diffusion). This novel model represents a significant advancement over traditional single-modal diffusion models. The MM-Diffusion model incorporates two coupled denoising autoencoders, each dedicated to audio and video respectively. These subnets are designed to gradually generate aligned audio-video pairs from Gaussian noises, a process distinctively different from existing methods. A key feature of this model is the introduction of a novel random-shift-based attention block, which efficiently bridges the audio and video subnets, ensuring semantic consistency across both modalities. The study showed results in both unconditional audio-video generation and zero-shot conditional tasks, like converting video to audio. This shows how flexible and useful the model is for multi-modal AI applications.

In conclusion, while the full spectrum of AGI-driven systems’ capabilities in education is yet to be fully explored, ongoing research and development in AGI provide a vital groundwork for future breakthroughs. As AI continues to evolve, it is poised to transform the educational landscape, offering more personalized, efficient, and immersive learning experiences through audio and video analysis and generation.

\section {Ethical, Explainable, and Responsible AGI in Education}\label{Lehong}
AGI represents an exciting frontier and a transformed force in education, given its potential to analyze and generate multimodal content across domains of cognition, creativity, and emotional intelligence, which positions AGI as a pivotal part of orchestrating highly personalized learning experiences. Nevertheless, with the recent advancement of LLMs, the rise of AGI in educational contexts introduces complex challenges and responsibilities. To harness AGI’s capabilities in facilitating multimodal learning while not compromising ethical standards, the implementation of educational AGI must be underpinned by a commitment to ethical integrity, system explainability and transparency, and responsible deployment through the efforts of researchers, educators, and AGI developers.
\subsection {Ethical Integrity in Educational AGI}

\subsubsection{The ethics of AI in general}
As with the advent of previous groundbreaking technologies, the growing ubiquity of artificial intelligence (AI) across all sectors of society has given rise to the emergence of complex ethical dilemmas and derived intense scholarly discussions. The discourse on AI ethics has been thoroughly examined in seminal works \citep{boddington2017towards, floridi2019translating, jobin2019aiethics, whittaker2018ai, winfield2018ethical}, highlighting the multifaceted nature of these ethical considerations. A critical aspect of this scholarly discourse pertains to data-related ethics, including the challenges of bias in datasets, the protection of data privacy, the imperative for unbiased assumptions and decision-making, and the pursuit of transparency throughout the processes of data collection, analytical processing, model development, and interpretive practices \citep{floridi2019translating, jobin2019aiethics, potgieter2020privacy}. Furthermore, an increasing discourse surrounding the multifaceted ethical landscape of AI is extending to the broader socio-economic impacts of AI, such as employment displacement and the need for workforce retraining \citep{acemoglu2018race}. This has led to the pressing need to establish fair labor practices that take into account the potential of AI to alter traditional employment structures, opportunity equity, and job security \citep{volini2020ethical}. 

AI's broader societal implications, such as its impact on social inequalities, necessitate inclusive design and diverse representation in AI systems \citep{benjamin2019race}. Encapsulating those concerns, \cite{morandin2023montreal} proposes a general human-centered framework for AI ethics, promoting principles such as autonomy, well-being, personal privacy, solidarity, democratic participation, responsibility, equity, diversity, prudence, and sustainable development. For example, high-stakes workforce domains, such as healthcare and criminal justice, confront significant ethical issues pertaining to AI's role in human decision-making, underscoring the importance of responsibility and transparency \citep{char2018implementing}.

\subsubsection{The ethics of AI in education}
Ethics have also received great attention in the field of Artificial Intelligence in Education (AIED). As with AI ethics in general, concerns about the large volumes of data collected for learning analysis, such as the recording of student performance data, the detected emotional states through facial expression and eye-tracking, is one critical focus of ethics in AIED \citep{holmes2021ethics}. Questions like who has the authority to access these data, what is the privacy protections, and who is responsible for the ethical use and interpretation of those data are broadly discussed in the AIED community. Furthermore, another major AI ethical concern within AIED centers on the learning analytical of student data. In this regard, it is imperative for researchers, educators, and policy-makers to critically consider questions, such as (1) How should the student data be analyzed, interpreted, and acted upon? (2) How should the biases of decision-making be prevented or ameliorated when considering gender, age, race, and family social status? 

Furthermore, the ethics of AIED accounts for the ethics of education, specifically beyond the above ethical questions about data or learning analytics. In this regard, the AIED research started to examine various issues that emerged or were enlarged by the integration of AI in educational settings. Those issues include, but are not limited to, (1) the access to education (e.g., educational fairness and equity to all students); (2) the roles of AI with respect to teachers (e.g., to replace or augment human roles); and (3) the purpose of using AI in learning (e.g., help the student pass exams or prepare for future AI-empowered workforce). In addition, the question about student agency \citep{holstein2019designing} is proposed when students, for example, follow AI tutors in personalized learning.

Concurrent with the incorporation of AI into educational settings, including AI-enhanced e-learning platforms and intelligent agents (such as intelligent tutoring systems and educational chatbots), a substantial body of research has been undertaken to explore the arising ethical considerations accordingly and specifically related to such AI-enhanced tools and learning environments. For example, researchers in AI-enhanced e-learning are interested in investigating ethical issues, such as user consent, equity and diversity, accessibility, identity and confidentiality, and student privacy (e.g., \cite{anwar2011facilitating, sacharidis2020fairness, akgun2021artificial}. For example, regarding student privacy, \cite{anwar2011facilitating} posited that "when an observer [e.g., an automated system] monitors students' behaviors with full knowledge of their identity and performance, the student being monitored does not enjoy any privacy" (p. 63). In addition, the ethical practices of students when they use AI-enhanced e-learning systems, such as cheating and plagiarism \citep{gearhart2012lack}, is another important scholarly focus. Similarly, researchers and technology developers of intelligent agents focus on ethical issues \citep{murtarelli2021conversation, richards2019supporting} for instance, transparency, privacy, and fairness. 

\subsubsection{The ethics of AGI in education}

AGI represents a new era and the next frontier in the evolution of AI; unlike specialized or narrow AI, which is adept at performing specific tasks, AGI endeavors to replicate comprehensive human cognitive capabilities, demonstrating versatility in understanding, learning, and adapting across various domains. The distinguished characteristic of AGI lies in its generalization capacity, adeptly transferring and applying knowledge and skills to new and unforeseen situations \citep{rayhan2023artificial, latif2023artificial}. This leap in capability and the widespread integration of pre-trained Large Language Models (LLMs)—some fine-tuned for particular domains—into various societal sectors, including education, has amplified the ethical conversation surrounding AI. Within AIED, the ethical concerns commonly addressed, such as bias and psudo AI bias \citep{RN3205}, fairness, privacy, transparency, and autonomy and accountability, are also critical in AGI's educational applications. Nevertheless, AGI's distinguished characteristics and increased potential to revolutionize education introduce additional ethical and legal considerations, such as those concerning the reproducibility and replicability of AGI models in specific domains and the corresponding human agency and autonomy. In their exploration of AI ethics in education and AGI ethics in particular, \cite{latif2023artificial} not only revisit concerns like bias, fairness, privacy, and trust in AIED but also introduce further considerations to AGI, such as model reproducibility and replicability, and displacement strategy. While \cite{latif2023artificial} provides a comprehensive technology-oriented analysis, emphasizing the emergent issues associated with AGI models and their capabilities, their discourse extends to the broader implications of AGI's ability to be reproduced, replicated, and its potential to displace human capabilities.

Expanding on the ethical concerns discussed by \cite{latif2023artificial}, this position paper delves deeper into the burgeoning ethical considerations associated with deploying AGI within education, adopting a more human-centered perspective. 

The issue of plagiarism is a long-discussed topic in the educational community. Plagiarism is defined by \cite{liaqat2011plagiarism} as "the act of imitating or copying others' creation or idea without permission and presenting it as one's own" (p.5). However, the burgeoning integration of AGI systems such as ChatGPT and Bing AI in educational settings has markedly intensified the ongoing discussion on its impact on academic integrity and the potential for student plagiarism when utilizing AGI on tasks such as writing an essay. Those concerns emerged due to the chatbot's ability to produce accurate and human-like responses, which has changed the dynamics of academic integrity \citep{currie2023academic}. In this context, \cite{slade2023academic} reviewed the key issues at the intersection of academic integrity and AI and discussed the role of large language models like OpenAI's GPT-3, which highlights the need for a novel understanding of academic integrity when using AGI. Therefore, it is critical to reevaluate the existing educational integrity guidelines and standards and call for innovative methods and techniques to detect and verify the originality of student works. In response to this call, \cite{cotton2023chatting, bissessar2023use} posited that various AGI tools, such as ChatGPT, not only offer benefits such as increased student engagement and collaboration but also raise concerns regarding academic integrity and plagiarism, shedding light on the benefits and challenges from both viewpoints. Accordingly, with the increasing adoption of LLMs and AGI by students for writing academic papers and completing various assignments, research on detecting student plagiarism suggests the critical for effective detection systems for AI-generated responses \citep{taloni2023modern,sahu2016plagiarism, francke2019potential}. 

Expanding on the ethical concerns regarding AGI-generated content in educational settings, the issue of AI credit and copyright when students use AGI tools for their assignments becomes increasingly complex. This ethical concern revolves around the proper attribution and appreciation of work created through AI, challenging traditional concepts of authorship and copyright. According to \cite{ZB2023@copyright}, the use of AGI in facilitating student assignments blurs the lines of intellectual property, which creates a gray area in defining authorship and originality and leads to potential conflicts in academic integrity policies. \cite{akgun2021artificial} calls for educational institutions to develop clear guidelines that delineate the acceptable use of AGI in completing academic assignments, ensuring that students understand the importance of original thought and effort in their academic pursuits. Additionally, \cite{foltynek2023enai} emphasizes the need for reevaluation of existing academic policies to incorporate the nuances of AGI-generated content and grant the copyright and credit to AI. To address the issue revolving around AI copyright and student work originality, we posit the importance of encouraging students to engage with AGI outputs critically, integrating them thoughtfully into their work with proper acknowledgment and copyright. 

The ethical issue of preserving human agency and autonomy in the context of the adoption of AGI tools in education is critical and increasingly discussed. The integration of AGI tools into educational practices brings forth the potential risk of diminishing human decision-making and creativity, a fundamental aspect of learning and cognitive development. \cite{prunkl2022human} posited that autonomy is a cornerstone of human dignity, making its preservation in the era of AI crucial. The potential of AGI to support and foster human autonomy is significant, yet there are substantial risks, such as AI-facilitated deception and manipulation, which could seriously interfere with human autonomy on a large scale. Moreover, the rapid development of AI, particularly AGI, poses significant educational challenges. The skills necessitated in various workforce areas will likely be replaced by AI automation, such as decision-making, critical thinking, creativity, teamwork, and communication \citep{levesque2018role}. Under this circumstance, K-12 schools and post-secondary institutions must prioritize those skills in education to maintain human agency and adapt to the changing workforce market. Meanwhile, when teachers implement AI tools in educational settings, they should carefully manage those advanced technologies to ensure they augment rather than replace human interaction and decision-making in the learning process \citep{levesque2018role}. In addressing the issue surrounding human agency and autonomy, it is critical to balance the benefits of AI innovation with the preservation of fundamental human values in shaping a future where technology serves as a tool/partner for enhancing human capabilities rather than supplanting them.

\subsection{Explainability and Transparency in Educational AGI}

The rapid advancement of AGI in educational settings highlights the critical need for enhanced explainability and transparency in AGI models and mechanisms, a concern increasingly discussed in the realm of AI ethics. This growing need for explainability and transparency in how AGI systems process and make decisions is increasingly recognized in the academic community, as evidenced by a body of research \citep{gade2019explainable, kumar2020explainable}. In educational settings, where AGI models are tasked with processing sensitive data, such as student responses, the implications of their predictions are profound, potentially influencing teachers' instructional practices \citep{susnjak2022learning}. In the domain of multimodel assessment, where AGI models are responsible for analyzing and generating various types of student responses -- including text, graphics, and video/audio -- the need for explainability and transparency becomes even more paramount. Teachers, who are the front-line implements of these technologies, must fully understand the rationale behind the AI's scoring and generative processes, as well as the criteria and algorithms it employs \citep{latif2023artificial, polak2022teachers}, to integrate them into their teaching strategies effectively. 

Moreover, the issue of bias in AI, particularly in educational settings, has been a topic of intense discussion, further emphasizing the need for transparent and explainable AGI models \citep{baker2021algorithmic}. Teachers need to be assured that the AI tools they use are effective and equitable in their treatment of student data.

Consequently, enhancing the explainability of AGI models, particularly those fine-tuned for specific academic disciplines \citep{latif2023fine}, transcends a mere technical challenge. It represents a fundamental ethical imperative, ensuring that the capabilities and multimodality of these models are aligned with and promote effective teaching practices. This alignment is crucial not just for educational efficacy but also for maintaining trust in AI technologies among educators and learners. Therefore, the development of AGI in education must consider these ethical imperatives, striving for models that are advanced in their capabilities and transparent, fair, and understandable to their human users \citep{yolcu2023redefining}.

\subsection{Responsible Use of Educational AGI}
\cite{dignum2019responsible} describes responsible AI as “about being responsible for the power that AI brings” (p.1), which is about ensuring that the results of the development and utilization of AI are beneficial for many. The main challenge of responsible AI is: who or what is responsible for actions and decisions made by AI systems \citep{dignum2019responsible}? AI systems, even with some level of human-like intelligence, are artifacts that can not be seen as responsible actors to take the responsibilities to determine and evaluate their behaviors and outcomes. 

In education, the question of where lies the responsibility, for example, for the identification of at-risk students or for the refusal of a college application, when these decisions are made by AGI systems or based on the results provided by such AI systems? Is the developer of the algorithm responsible, the data providers, or the users who accepted AGI's decisions? Answering those questions and distributing responsibility appropriately presents a complex challenge, necessitating a shared approach to responsibility that engages various stakeholders, such as AI developers, educators, policymakers, and researchers, through various stages of AI system development and application. 

Furthermore, LLMs, machine learning algorithms capable of recognizing and producing high-volume human-like texts, are being increasingly used by learners to create content that is often indistinguishable from human-written information and, in some instances, includes compelling but misleading information \citep{kreps2022all,zhou2023synthetic}. For example, Galactics, an LLM for science that can summarize academic papers, solve math problems, and generate Wiki articles, was found to generate biased and even incorrect results, such as generating fake papers and attributing references to real researchers \citep{Heaven2022meta}, thus was taken down soon after its release by Meta. Likewise, as ChatGPT further brought LLMs into public sections, including education, it was criticized for biased or false outputs \citep{Birhane2022gpt,wach2023dark}. The deployment of LLMs and AGI for generating misinformation is detrimental to learners in their accurate understanding of domain knowledge and concepts, especially considering that many lack the proactive abilities and awareness necessary to identify AGI-generated misinformation. Educators are responsible for equipping learners with the critical thinking skills and digital literacy necessary to discern the credibility of AGI-generated content. For instance, educators can encourage learners’ skepticism and the questioning of sources provided by AGI, as well as provide instruction on evaluating the trustworthiness of AGI-generated content. The critical responsibility lies with humans rather than AI systems. Furthermore, scholarly research has also focused on developing theoretical and practical methods for identifying AI-generated misinformation. In a recent study, \cite{zhou2023synthetic} investigated the distinct features of AI-generated misinformation by comparing them with human-created misinformation and introduced a theory-guided technique to detect such content effectively. 

In addition, the integration of AGI into educational contexts has been mixed with enthusiasm and concerns. Given the unreliability of AI detectors in identifying AI-generated content and their potential risks of violating student data privacy \citep{chaka2023detecting,sankar2023can}, it raises an important question: What actions can educational institutions take, aside from relying on detectors, to guide the appropriate integration of generative AI within their contexts? First, instructors and students can benefit from training workshops, practical seminars, and open discussions that aim to provide a foundational understanding of AI \citep{iqbal2022exploring}. These events can deepen the understanding of AI's technical functions, opportunities, and potential risks and enrich teachers’ and students’ general awareness of its impact. Second, schools need to offer syllabus templates and/or samples that are tailored to the specific cultures and needs of various departments and programs. This helps instructors to make their own policy decisions and promote transparency in their teaching \citep{wang2023seeing}. Third, in order to prepare instructors for class design, schools can develop and provide a range of pedagogical strategies, which may include using AI to enhance teaching and learning along with some preventive measures against its potential misuse \citep{rosyanafi2023dark}. Last but not least, AI professionals and institutions’ teaching and learning centers can also offer one-on-one consultation services. These would be valuable resources that can support educators and students in addressing their specific concerns and opportunities related to AI in their own teaching and learning contexts.

\section {Discussion}\label{}

While the field is embracing the releases of Gemini \citep{deepmind_gemini} and GPT-4V\citep{GPT-4V}, we present this comprehensive examination of multimodal AI's role in advancing AGI within educational contexts. It critically assesses the evolution of AI in education, emphasizing the importance of multimodality in learning, which encompasses auditory, visual, kinesthetic, and linguistic modes. The study delves into key facets of AGI, including cognitive frameworks, knowledge representation, adaptive learning, strategic planning, natural language processing, and multimodal data integration \citep{zhai2023can}. This extended introduction should explore the implications of these facets for the future of education, considering the dynamic nature of learning environments and the increasing demand for personalized education. As the study highlights, integrating different learning modes can lead to more effective and personalized educational experiences. This sets a new direction for future research and practical applications of AGI in education.

Unveiling human cognitive processes provides a multimodal framework for AGI, underscoring the necessity of mimicking human cognitive processes for effective AGI systems for education. Prior research has revealed that students' learning can benefit from multimodalities and the generative and classifying functions of AGI can potentially contribute to student learning \citep{sweller2011cognitive,GINNS2005313,
moreno2002learning}. This aligns with and extends existing literature, emphasizing the importance of reasoning, learning, memory, and perception in AGI development to assist in human learning and problem-solving. It contributes to the field by highlighting the need for AGI-driven systems to model the structural and functional principles of the human mind, a perspective that enhances understanding of AGI's potential in educational settings.

Insights of this study into knowledge representation and manipulation highlight AGI's capability to accumulate, apply, and update knowledge based on new experiences. Students come into the classroom with prior knowledge and information channel preferences. Without considering these differences, it is challenging for educators to provide customized learning support, addressing equity and diversity\citep{miller2021supporting}. Our findings support the development of AGI-driven systems for effective and equitable learning. The insights are consistent with existing research but expand it by incorporating techniques like logical reasoning, probabilistic models, and knowledge graphs. The paper contributes to the literature by a detailed examination of these techniques and their applicability in educational AGI systems, offering new perspectives on enhancing AGI's decision-making and problem-solving abilities for learning.

The research emphasizes AGI's need for flexibility and experience-based adaptation to support multimodal learning, utilizing machine learning methods like deep reinforcement learning. AGI, with adaptability, could meet students's needs regardless of their backgrounds, levels of learning, or special needs \citep{Panjwani2023}. This finding adds to the literature by demonstrating how these learning methods leverage AGI to meet individual needs and improve students' performance. This contribution is significant for its practical implications in education, suggesting ways AGI can be tailored to various learning styles and environments.

The findings on AGI's capabilities in strategic planning and decision-making, involving search algorithms and game theory, present a novel contribution. While AGI cannot substitute teachers, it can assist and collaborate with teachers to implement effective instruction by assisting in planning and instructional decision-making. Aligning with some existing theories, this aspect of the paper underscores AGI's potential in tackling complex tasks and challenges in educational settings. It broadens the current understanding of AGI's applicability, suggesting its potential role in educational strategies, pedagogy, and curriculum planning.

To sum up, the paper adds substantially to the literature on AI in education, particularly in exploring AGI. By focusing on multimodality, it not only aligns with existing research but also extends it, providing new insights into how AGI can be effectively integrated into educational contexts. The study's contributions lay the groundwork for future research and practical applications of AGI in education, highlighting its transformative potential.

\section {Conclusions}\label{}
This paper has explored the pivotal role of multimodal AI in steering the progress toward AGI within educational contexts. Our investigation has underscored the critical integration of diverse sensory channels and cognitive strategies, highlighting how multimodal AI complements and enhances the educational experience. This integration facilitates a more comprehensive, inclusive, and effective learning process, accommodating various learning styles and needs.

The transformative potential of AGI in education is profound. As we venture into the realm of AGI, the traditional teaching and learning paradigms are set to undergo significant transformations. AGI's capabilities in providing personalized, dynamic, and interactive learning experiences promise to revolutionize educational methodologies, making them more adaptive and responsive to individual learner's requirements. This evolution is crucial for addressing a global learning population's diverse and ever-changing educational needs. Our findings emphasize the necessity of continued research in this domain. As AGI continues to evolve, exploring and understanding its implications is imperative. This includes technological advancements and the ethical, social, and practical aspects of integrating AGI into educational systems.

In conclusion, the journey towards fully realizing the potential of AGI in education is both challenging and exhilarating. It calls for collaborative efforts from educators, researchers, technologists, and policymakers. The broad, transformative impact of AGI in reshaping educational landscapes holds the promise of a future where education is more accessible, engaging, and effective for learners across the globe. Thus, our exploration into the multimodal dimensions of AI for education sets a foundational framework for future developments in this exciting and rapidly evolving field.

\section*{Acknowledgement}
This presentation was funded by the National Science Foundation(NSF) (Award no. 2101104, 2138854). Any opinions, findings, conclusions, or recommendations expressed in this material are those of the author(s) and do not necessarily reflect the views of the NSF.

\section*{Note. Pioneering AGI Integration in Education: UGA's Initiatives}
To integrate AGI into classroom and undergraduate education, the University of Georgia (UGA) has launched several notable initiatives. A prime example is UGA's Generative AI initiative (\url{https://ai.uga.edu/genai}). This initiative is at the forefront of exploring and harnessing the potential of generative AI models within educational contexts. It serves as a platform for investigating how these advanced AI technologies can be applied across various educational settings. 
%Professor Xiaoming Zhai is a key contributor to this initiative, 
Interdisciplinary experts from computer science, STEM education, educational psychology, educational technology are key contributors to this initiative, bringing a wealth of expertise and insight to the forefront of this exploratory endeavor.

In addition to the Generative AI initiative, the AI4STEM Education Center at UGA (\url{https://ai4stem.uga.edu/}) represents a significant step towards the integration of AGI into STEM education. 
%Under the leadership of Professor Zhai as the Director and 
With a dedicated interdisciplinary team, the AI4STEM Education Center focuses on the application of AI tools and methodologies to enhance teaching and learning in STEM fields. The center is committed to creating an educational landscape where AGI is utilized to innovate and transform traditional teaching and learning practices. Together, these initiatives at UGA highlight a proactive and pioneering approach to embedding AGI technologies in educational frameworks, setting an example for academic institutions globally.

\bibliographystyle{ACM-Reference-Format}
\bibliography{sample-base}

\end{document}